\documentclass[times,twocolumn,final]{elsarticle}

\usepackage{ipcai_arxiv}
\usepackage{framed,multirow}

\usepackage{amssymb}
\usepackage{latexsym}

\usepackage{url}
\usepackage{cite}
\usepackage{amsmath,amssymb,amsfonts,bm}
\usepackage{booktabs}
\usepackage{arydshln}
\usepackage{tabulary}
\usepackage{tabularx}
\usepackage{multirow}
\usepackage{graphicx,epstopdf}
\usepackage{textcomp}
\usepackage{subcaption}
\usepackage{booktabs}
\usepackage[svgnames,table]{xcolor}
\usepackage{pifont}
\usepackage{arydshln}
\usepackage[colorlinks=true,citecolor=cyan,urlcolor=cyan]{hyperref}
\usepackage{graphicx}
\usepackage{placeins}
\usepackage{fancyhdr}
\usepackage{float}
\newcommand{\xmark}{\ding{55}}
\fancypagestyle{firstpagestyle}{
    \fancyhf{} 
    \fancyhead{} 
    \fancyfoot{} 
}

\begin{document}


\title{Multi-view Video-Pose Pretraining for Operating Room Surgical Activity Recognition}

\author[1]{Idris \snm{Hamoud}\corref{cor}}
\author[1,2]{Vinkle \snm{Srivastav}}

\author[3]{Muhammad Abdullah \snm{Jamal}}

\author[2,4]{Didier \snm{Mutter}}

\author[3]{Omid \snm{Mohareri}}

\author[1,2]{Nicolas \snm{Padoy}}

\cortext[cor]{Corresponding author: ihamoud@unistra.fr}

\address[1]{University of Strasbourg, CNRS, INSERM, ICube, UMR7357, France}

\address[2]{IHU Strasbourg, Strasbourg 67000, France}

\address[3]{Intuitive Surgical Inc., Sunnyvale, USA}

\address[4]{University Hospital of Strasbourg, Strasbourg 67000, France}

\received{XXX}
\finalform{XXX}
\accepted{XXX}
\availableonline{XXX}
\communicated{XXX}

\begin{abstract}

\textbf{Purpose: } Understanding the workflow of surgical procedures in complex operating rooms requires a deep understanding of the interactions between clinicians and their environment. Surgical activity recognition (SAR) is a key computer vision task that detects activities or phases from multi-view camera recordings. Existing SAR models often fail to account for fine-grained clinician movements and multi-view knowledge, or they require calibrated multi-view camera setups and advanced point-cloud processing to obtain better results. 

\textbf{Methods: } In this work, we propose a novel calibration-free multi-view multi-modal pretraining framework, called \textit{Multiview Pretraining for Video-Pose Surgical Activity Recognition} (\textit{PreViPS}), which aligns 2D pose and vision embeddings across camera views. Our model follows CLIP-style dual-encoder architecture: one encoder processes visual features, while the other encodes human pose embeddings. To handle the continuous 2D human pose coordinates, we introduce a \emph{tokenized discrete representation} to convert the continuous 2D pose coordinates into discrete pose embeddings, thereby enabling efficient integration within the dual-encoder framework. To bridge the gap between these two modalities, we propose several pretraining objectives by using \emph{cross- and in-modality geometric constraints} within the embedding space and incorporating \emph{masked pose token prediction} strategy to enhance representation learning. 

\textbf{Results: } Extensive experiments and ablation studies demonstrate improvements over the strong baselines, while data-efficiency experiments on two distinct operating room datasets further highlight the effectiveness of our approach. We highlight the benefits of our approach for surgical activity recognition in both multi-view and single-view settings, showcasing its practical applicability in complex surgical environments. Code will be made available at \url{https://github.com/CAMMA-public/PreViPS}.

\end{abstract}

\maketitle
\thispagestyle{firstpagestyle}

\section{Introduction}
\label{sec:introduction}
The modern Operating Room (OR) is a high-stakes, fast-paced socio-technical environment where clinicians work collaboratively to ensure safe and efficient surgical procedures. To support these efforts, ORs are increasingly equipped with advanced sensors, including external cameras, to monitor and analyze clinical activities. By leveraging these sensor-enhanced capabilities of the OR, context-aware systems have emerged as a promising tool to optimize clinical workflows, support intra-operative decision-making, and enable early detection of adverse events through automated analysis of clinical processes~\citep{maier2022surgical}. Recent developments in OR applications highlight this potential, including radiation risk monitoring in hybrid surgeries~\citep{Rodas2018AugmentedRF,Ladikos2010EstimatingRE}, surgical workflow recognition~\citep{Padoy2008OnlineRO,Czempiel2020TeCNOSP}, and semantic scene understanding~\citep{LatGraphAdit,Koksal2024SANGRIASV}.

A key component of such systems is \emph{surgical activity recognition} (SAR), which aims to detect different activities or phases in long untrimmed videos recorded from external multi-view cameras. Recent SAR models~\citep{zhang2021real,Twinanda2017MultiStreamDA,Sharghi2020}, inspired by advances in action recognition, use clip-based approaches to segment videos into temporal phases. However, these approaches do not fully exploit the multi-view knowledge from the multi-camera setups and mainly rely on clip-level or global image features, overlooking fine-grained details of clinicians' movements. Some recently proposed methods based on the 4D-OR dataset address this limitation but require calibrated multi-view camera systems and advanced point-cloud processing for semantic scene graph generation, which is then used for surgical activity recognition~\citep{Ege_IJCARS}. However, these methods can be computationally expensive and rely on calibrated multi-view camera setups. These are challenging to acquire in practical OR settings, especially in robot-assisted surgical procedures where vision cameras are mounted on the movable surgical robot~\citep{Sharghi2020}.

As clinicians are the main dynamic actors in the OR, their fine-grained localization is crucial for reliable SAR systems. Human pose estimation, a computer vision task that localizes 2D body keypoints, has started to work remarkably well even in complex scenarios~\citep{cao2017realtime,PCT}. By explicitly integrating fine-grained pose information, SAR models can achieve significant improvements in activity recognition accuracy.

In parallel, computer vision has been witnessing significant advances in multi-modal pretraining~\citep{CLIP,DeCLIP,jia2021scaling} - a paradigm that bridges vision and language modalities. Models like CLIP~\citep{CLIP} and ALIGN~\citep{jia2021scaling} have demonstrated the ability to learn generalized multi-modal representations by aligning visual concepts with natural language descriptions using large-scale paired image-caption datasets. These models have enabled a shift from task-specific to more generalist models in a unified framework capable of handling diverse downstream tasks~\citep{zou2024segment,lin2023match}.

Motivated by these developments, this work introduces and investigates a key research question: how can human pose representations be effectively aligned with \emph{uncalibrated} multi-view camera images in a \emph{multi-view multi-modal pretraining} framework? By addressing this question, we aim to improve the performance of SAR systems as a downstream task by leveraging human pose estimation, multi-modal pretraining, and multi-view video analysis.

However, the task is non-trivial and presents challenges regarding suitable architecture design and effective pretraining objectives. From an architectural perspective, we propose a dual-encoder that processes both vision and human pose modalities, similar to common vision-language architectures~\citep{CLIP}. However, unlike vision-language architectures where text is a \emph{discrete} modality - with words or subwords transformed into discrete token representations - the human pose is typically represented as \emph{continuous} 2D keypoints. To overcome this challenge, we propose to use the \emph{Pose as Compositional Tokens} (PCT)~\citep{PCT}, which tokenizes the continuous 2D human pose coordinates into discrete tokens. These tokenized embeddings convert the continuous poses into discrete tokens and handle occlusions well by leveraging the dependency between joints encoded in the discrete pose tokens.

Building on this architecture, we design a set of \emph{pretraining objectives} to align pose and vision embeddings while exploiting the multi-view context. The pretraining objectives follow the concept of CLIP~\citep{CLIP}, where discrete pose embeddings are brought closer to the corresponding view's image embeddings, and embeddings of different images are pushed apart using InfoNCE loss~\citep{oord2018representation}. In the multi-view setting, we propose to extend further the idea to achieve \emph{view invariance}: the model not only brings the pose embedding closer to its corresponding camera view but also aligns it with embeddings from other camera views at the same time stamp, and vice versa.

While these constraints help align multi-view pose and vision embeddings, they may still lack geometric alignment, leading to suboptimal downstream performance. To address this, we propose two additional geometric constraints to improve representation quality: \emph{cross-modality constraints} -  these constraints ensure that pose and visual embeddings are geometrically consistent across modalities, and \emph{in-modality constraints} - these constraints enforce consistency within the pose or visual modality itself, enhancing structural coherence, similar to~\citep{Goel2022CyCLIPCC}.

Finally, we also leverage \emph{masked modeling}, a technique widely used in visual and language representation learning~\citep{MAE,VideoMAE,MaskFeat}. In masked image modeling, a portion of an image is hidden, and the model learns to predict the masked content based on its surroundings. Instead of applying this at the pixel level, we extend the idea to \emph{pose tokens}. Specifically, we mask a subset of pose tokens and feed them into a transformer-based backbone, which learns to output a feature representation of the masked content. These representations are then input to a transformer decoder to predict the missing pose coordinates, encouraging the model to learn a robust representation of pose information.

In summary, this work introduces a novel \emph{multi-view, multi-modal pretraining framework} by incorporating pose as compositional tokens, aligning embeddings across uncalibrated camera views, enforcing geometric constraints, and leveraging masked pose token prediction. We evaluate our framework on the SAR downstream task, conducting extensive ablation studies to analyze the contributions of each component and their impact on overall performance. A key highlight of our approach is its adaptability: even when finetuned with a single modality, our multi-modal pretraining framework achieves significant performance gains. Overall, we achieve significantly better results against strong baselines, thereby pushing the boundaries of surgical activity recognition, enabling a more accurate and reliable understanding of clinical workflows in calibration-free multi-view camera setups.

\section{Related work}
Our work relates to three broad areas of research: (a) data-enabled surgical video understanding, (b) skeleton-based action recognition methods, and (c) multi-modal representation learning for efficient utilization of multiple modalities through unsupervised pretraining objectives. 

\noindent \textbf{Surgical Video Understanding from External Cameras: } Research on OR workflow understanding has evolved across several key areas, including human pose estimation for clinicians~\citep{MVORVinkle,srivastav2020self}, radiation risk monitoring during surgical interventions~\citep{rodas2016see,Ladikos2010EstimatingRE}, and surgical workflow recognition~\citep{zhang2021real,Twinanda2017MultiStreamDA,MoreThan}. The OR-AR dataset was introduced as a multi-view dataset designed for robotic-assisted surgery (RAS) for surgical activity recognition~\citep{Sharghi2020}. This dataset has enabled the development of various SAR methods, including 3D CNNs with two-stage learning~\citep{Sharghi2020}, object-centric representation learning~\citep{hamoud24a}, and multi-modal self-supervised learning using intensity and depth images~\citep{Jamal-ISI}.

Another benchmark dataset, called the 4D-OR dataset~\citep{4DOR}, was introduced with densely annotated 3D scene graphs derived from simulated surgical procedures in an orthopedic OR. The 3D scene graph modeling of the OR has enabled the identification of clinicians' roles and OR activities through geometric and semantic interactions~\citep{4DOR}. Expanding on this, the LABRADOR framework incorporated a scene graph memory queue to leverage temporal dependencies in scene graph generation, demonstrating strong performance in activity recognition using heuristics derived from the generated scene graphs~\citep{Labrador,Ege_IJCARS}.

\begin{figure*}[t!]
    \centering
    \includegraphics[width=\textwidth]{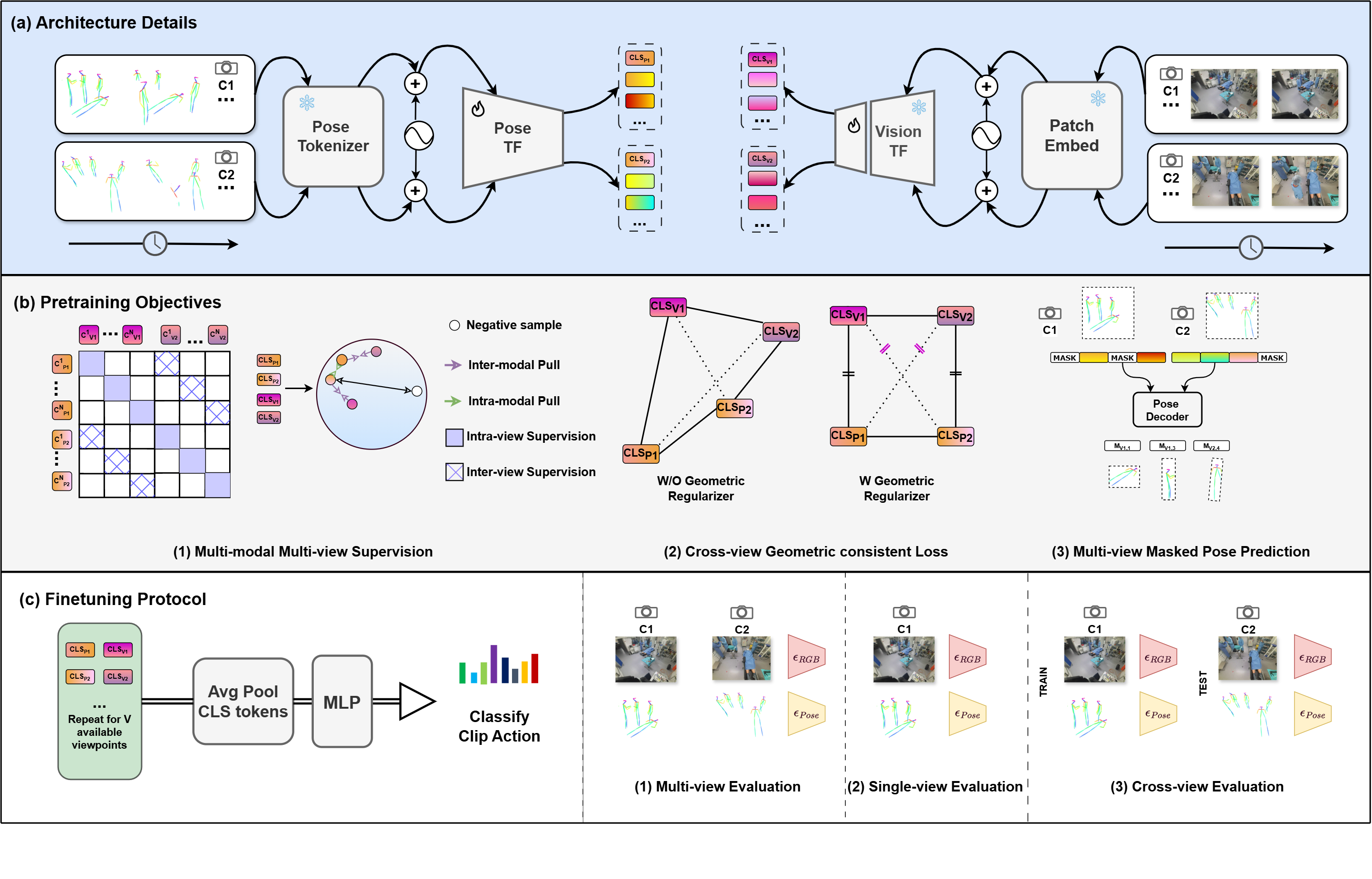}
    \caption{\textbf{Overview of our framework:} (a) Given a video clip, we first extract all human poses using ViTPose-Base~\citep{vitpose}. We tokenize the poses using PCT~\citep{PCT} and use a two-stream approach with MaskFeat~\citep{MaskFeat} on the vision features. (b) We use different pretraining objectives on the global representations of each modality and viewpoint. (c) We present our finetuning protocol, utilizing global representations from various modalities and viewpoints. Additionally, we demonstrate the versatility of our approach, enabling us to train and test our methods using different viewpoints.}
    \label{fig:enter-label}
\end{figure*}

\noindent \textbf{Skeleton-based Action Recognition: } Understanding human pose information and its dynamics is fundamental for human action recognition, as highlighted by foundational work in~\citep{Johansson1973VisualPO}. With significant advancements in human pose estimation~\citep{cao2017realtime,PCT}, skeleton-based action recognition has become a prominent area of exploration in computer vision~\citep{duan2022revisiting,zhou2023learning}. To mitigate background bias in action recognition, fine-grained human activity datasets such as NTU RGBD~\citep{shahroudy2016ntu} and Toyota Smarthome Untrimmed~\citep{Dai2020ToyotaSU} have been developed, providing more robust benchmarks for evaluation. Advancements in this field include the development of a hierarchical recurrent neural network (RNN) model, which partitions the human skeleton into subparts to capture long-term contextual information~\citep{Du_RNN}. Further progress was made with spatial-temporal graph convolutional networks (STGCNs), which model skeleton joints as a space-time graph to simultaneously capture spatial and temporal dependencies~\citep{kipf2016semi,STGCN}. While these methods have achieved strong results, they primarily depend on accurate 3D human poses obtained from motion capture systems~\citep{MoCAP}. This reliance limits their applicability in real-world environments, such as ORs, where poses are often noisy and of lower quality.

\noindent \textbf{Multi-view Representation Learning}

\noindent \textbf{Unimodal Pretraining:} Contrastive learning is widely used in self-supervised learning and has achieved strong results in vision tasks~\citep{SimCLR,MoCo,Swav,BYOL}. It ensures that positive sample pairs are close together while negative pairs are pushed apart. 
SimCLR~\citep{SimCLR} used large mini-batch sizes for diverse negative samples. MoCo~\citep{MoCo} addressed the computational bottleneck with momentum encoders and a sample queue, while SwAV~\citep{Swav} and BYOL~\citep{BYOL} opted to eliminate negative samples entirely.


In parallel, generative methods inspired by natural language processing representation learning have been adapted for vision tasks. Models like BeIT~\citep{BEiT}, MAE~\citep{MAE}, and MaskFeat~\citep{MaskFeat} use a masking strategy to hide parts of an image and train the model to predict them. VideoMAE~\citep{VMAE} extends this approach to video data by adding a temporal dimension to the masking process. 

\noindent \textbf{Multi-modal Pretraining:} Recent multi-modal pretraining approaches leverage natural language supervision to effectively transfer knowledge to various downstream tasks. Methods like CLIP~\citep{CLIP} and ALIGN~\citep{jia2021scaling} use contrastive learning~\citep{oord2018representation} to align images with their corresponding captions, enhancing the robustness of learned representations. Building on this, AlBeF~\citep{AlignBF} introduces a momentum encoder, inspired by MoCo~\citep{MoCo}, to mine hard negative samples and address computational bottlenecks in CLIP-like methods. A common challenge in multi-modal models is \emph{modality laziness}, where one modality dominates, leading to suboptimal performance~\citep{modLaz}. To mitigate this, recent work integrates local structural information into multi-modal learning, improving robustness and addressing modality imbalance~\citep{Yang2022VisionLanguagePW}. On the generative side, encoder-decoder models have been explored for multi-modal pretraining. For example, ImageBeIT~\citep{ImageBeIT} adapts the BeIT framework~\citep{BEiT} to incorporate both image and caption signals for cross-modal reconstruction, as demonstrated in~\citep{ImageBeIT,PeVLPV}.

In contrast to these approaches, our framework focuses on learning multi-modal representations using human pose and appearance modalities. A key distinction lies in the design of suitable pretraining objectives to enforce viewpoint consistency across camera views, thereby enhancing the multi-modal representations and subsequently improving surgical activity recognition performance.

\section{Methods}\label{method}

In this section, we introduce \emph{PreViPS}, our calibration-free multi-view multi-modal pretraining framework for surgical activity recognition (SAR). We first introduce the problem and then describe the \emph{dual-encoder architecture}, detailing the \emph{video encoder} for extracting visual features and our novel \emph{pose encoder}, which converts continuous 2D human pose coordinates into discrete embeddings. Next, we describe our three multi-view multi-modal pretraining objectives, which align video and pose embeddings across camera views by enforcing \emph{cross- and in-modality geometric constraints} and leveraging \emph{masked pose token prediction}. Finally, we explain the \emph{model finetuning process} for downstream SAR tasks, optimizing the learned representations for both multi-view and single-view surgical activity recognition. Through these architectural and training choices, \emph{PreViPS} enables robust and efficient activity recognition in complex surgical environments.  

\subsection{Problem Overview}

Given a training dataset of multi-view human-centric video clips $\mathcal{D} = \left\{ \mathbf{x}| \mathbf{y^*} \right\}$ where $ \mathbf{x} \in \mathbb{R}^{C \times T \times 3 \times H \times W}$ is a multi-view video-clip set captured by $C$ cameras over $T$ frames with resolution $H \times W$, and $\mathbf{y^*} \in \mathbb{R}^{C \times  T \times N_{p} \times 2 \times N_j}$ represents the pseudo 2d human poses for $N_{p}$ persons with $N_{j}(=17)$ number of joints (body keypoints) generated by an off-the-shelf human pose estimator, the goal to learn a joint latent space that correlates semantically similar video clips with the corresponding poses across camera views and vice versa. 

Formally, our goal is to learn two mappings: $\mathcal{F}: \mathbf{x} \rightarrow \mathbb{R}^{C \times D}$ and $\mathcal{G}: \mathbf{y^*} \rightarrow \mathbb{R}^{C \times D}$, which map a video clip and 2d pseudo poses into a $C \times D$ dimensional latent vector, where \( D \) represents the embedding dimension. To learn these two mappings, we employ a dual-branch model with a vision branch $\mathcal{F}$ and a pose branch $\mathcal{G}$ using transformer-based \emph{dual-encoder architecture}, described as follows.

\subsection{Dual-encoder Architecture}

\subsubsection{Video Encoder} 
We employ \emph{MaskFeat}~\citep{MaskFeat} as our video encoder. Given a multi-view video clip from \( C \) camera viewpoints, the encoder first applies a \emph{patch embedding} layer, which employs convolution and linear projection to transform the video clip into a sequence of tokens. These token sequences are then processed by a Vision Transformer to produce contextualized video embeddings \( I_c \in \mathbb{R}^D \) for each camera \( c \in [1, \dots, C] \), where \( I_c \) corresponds to the special [CLS] token, referred to as $CLS_{I}^{c}$, used in Vision Transformers.


\subsubsection{Pose Encoder}
\label{pose_encoder}

\paragraph{Pose Token Representation}
Given a video clip, we use the VitPose-B~\citep{vitpose} as an off-the-shelf pose estimator to generate pose sequences for each camera view. We also gather identity information for each detected pose using the established SORT~\citep{Bewley2016SimpleOA} algorithm.

Let $ p_{i,t}^{c} \in \mathbb{R}^{2 \times N_{j}}$ be the acquired pose coordinates at camera viewpoint $c \in [ 1,...,C]$ for a person $i$ and timestep $t$. To obtain a compact and meaningful representation for each single human pose, we pass $p_{i,t}^{c}$ through a frozen pose tokenizer~\citep{PCT} to generate the following bottleneck representation: $\pi_{i,t}^{c} \in \mathbb{R}^{D}$. This structured representation models the dependency between body joints and provides a distinct discrete representation similar to the text modality in vision-language pretraining. For each camera stream, we also append a learned vector [CLS]  as the first token of each sequence and use the output vector corresponding to that position for clip-level action recognition. More specifically, each camera stream pose latent representation is represented as,
\begin{equation}
    \mathcal{Y}^{c} = \{CLS_{J}^{c}, \pi_{1,1}^{c}, ..., \pi_{N_{p},T}^{c}\} \in \mathbb{R}^{D \times (N_{p} \times T + 1)}
\end{equation}

In this notation, $N_{p}=8$ is the maximum number of detected persons per frame. To ensure a consistent number of inputs, if the number of clinicians in a frame is less than $N_{p}$, we pad the sequence with a special $PAD$ token.

\paragraph{Positional Embeddings}
To encode spatiotemporal information in the pose sequences, we incorporate positional embeddings for various attributes such as time, track ID for persons, and viewpoint ID. 

Concerning viewpoints, we adopt the method proposed by Geng et al.~\citep{Geng}, which involves introducing learnable 1D parameters that represent each viewpoint and timestep. For time and track ID, we utilize 2D sine and cosine functions as a form of positional encoding. These parameters are added to the features of each video pose token captured from different perspectives. 

\paragraph{Network Architecture}
Given the previously defined representation, we adopt a vanilla transformer~\citep{transformer} as the backbone network. The pose embeddings (aggregated with positional embeddings) described above are fed to the pose transformer $\mathcal{M}$.
\begin{equation}
    \widehat{\mathcal{Y}} = \mathcal{M}(\Theta,\mathcal{Y})
\end{equation}
Here, $\Theta$ is the model parameters, and $\widehat{\mathcal{Y}}$ is the updated latent representation for pose information.
The pose transformer comprises a stack of ${L = 6}$ multihead self-attention layers. Each layer in the pose transformer $\mathcal{M}$ has a standard architecture consisting of multi-head self-attention modules and feed-forward networks. The encoder outputs a sequence of pose embeddings of dimension ${D}$.

\subsection{Aligning video and pose embeddings}

In this section, we outline our approach to cross-modal alignment. Our contrastive objective is to optimize the encoders specific to each modality. These encoders map the global embeddings from each modality and viewpoint to ensure their representations are closely aligned. Let $I^{c}=CLS_{I}^{c}$ represent the learned embeddings for the video modality, and $J^{c} = CLS_{J}^{c}$ represent the learned embeddings for the pose modality, both at each camera viewpoint $c$.

\subsubsection{Multi-modal Contrastive Learning}
\label{multi-modalContrast}
\paragraph{Cross-Modality Alignment}
Let us first address the cross-modal alignment between image and pose modalities. We want embeddings of the same samples from two viewpoints to be close to each other. Thus for each pair of camera views $(p,q) \in [1,...,C]^{2}$, we aim to bring $I_{n}^{p}$ and $J_{n}^{q}$ closer together while pushing apart the other embeddings from the remaining samples in the minibatch of batch size $N$.

\begin{equation}
    \mathcal{L}_{I/J}= -\dfrac{1}{N} \sum_{n=1}^{N}\sum_{(p,q)\leq C}\log(\frac{\exp(\langle I_{n}^{p}, J_{n}^{q} \rangle / \tau)}{\sum_{k=1}^{N} \exp(\langle I_{k}^{p}, J_{k}^{q} \rangle/ \tau)}) \label{eq:1}
\end{equation}
Here $\tau$ is the temperature hyper-parameter that regulates the penalty to the hard negative samples.
\paragraph{In-Modality Alignment}
Similar to cross-modal alignment, we also propose objectives to increase the similarity of embeddings from the same modality that come from different viewpoints.
\begin{equation}
    \mathcal{L}_{I/I}= -\dfrac{1}{N} \sum_{n=1}^{N}\sum_{(p,q)\leq C}\log(\frac{\exp(\langle I_{n}^{p}, I_{n}^{q} \rangle / \tau)}{\sum_{k=1}^{N} \exp(\langle I_{k}^{p}, I_{k}^{q} \rangle/ \tau)}) \label{eq:2}
\end{equation}

We can define the $\mathcal{L}_{J/I}$ and $\mathcal{L}_{J/J}$ losses reflexively by adjusting the embeddings in the loss accordingly. The multi-modal contrastive objective in PreViPS aims to align the video and pose representations by minimizing the loss function $L_{Con}$ defined below:
\begin{equation}
    \mathcal{L}_{Con}= \dfrac{1}{4} (\mathcal{L}_{I/J} + \mathcal{L}_{J/I} + \mathcal{L}_{I/I} + \mathcal{L}_{J/J})
\end{equation}

We refer to the pretraining following this loss, $\mathcal{L}_{Con}$, as CLIP* as it is an adaptation of the CLIP objectives to the video-pose modalities with multi-view constraints.
\paragraph{Sampling Policy}
Equations~\eqref{eq:1} and~\eqref{eq:2}, are computed over $N$ training instances, each in the form of a video-clip pose pair. A naive sampling policy may randomly sample instances from adjacent temporal segments, leading to semantically similar negative samples. This may confuse the model and hurt the final downstream performance. Therefore, we force each instance of our minibatch to be temporally distant. We divide the complete video into $N$ segments, where $N$ is the batch size, and sample one instance from each segment.

\subsubsection{Geometric Consistency}
We propose to incorporate geometric constraints into our pretraining objectives, similar to CyCLIP in vision-language pretraining~\citep{Goel2022CyCLIPCC}. We aim to mitigate inconsistencies in the shared embedding spaces of video and pose representations across different viewpoints. To achieve this, we introduce two geometric consistency regularizers, which are defined over each mini-batch as follows:

(1) \emph{Cross-Modal Geometric Consistency Loss}: This loss minimizes discrepancies in similarity scores for video-pose pairs across different viewpoints. It is formulated as:
\begin{equation}
    \begin{aligned}        \mathcal{L}_{C-Geo}= \dfrac{1}{N} \sum_{n=1}^{N}\sum_{(p,q)\leq V}&(\langle I_{n}^{p}, J_{n}^{q} \rangle - \langle J_{n}^{p}, I_{n}^{q} \rangle)^{2}, 
    \end{aligned}
\end{equation}
where \(I_{n}^{p}\) and \(J_{n}^{q}\) represent video and pose embeddings, respectively, for viewpoint \(p\) and \(q\). Also, \(\langle I_{n}^{p}, J_{n}^{q} \rangle\) and \(\langle J_{n}^{p}, I_{n}^{q} \rangle\) measure the similarity between video and pose embeddings for different viewpoints.

(2) \emph{In-Modal Geometric Consistency Loss}: This loss ensures that similarity scores remain consistent across viewpoints across video and pose pairs. It is defined as:
 \begin{equation}
    \begin{aligned}        \mathcal{L}_{I-Geo}= \dfrac{1}{N} \sum_{n=1}^{N}\sum_{(p,q)\leq V}&(\langle I_{n}^{p}, I_{n}^{q} \rangle - \langle J_{n}^{p}, J_{n}^{q} \rangle)^{2}, 
    \end{aligned}
\end{equation}
Here, \(\langle I_{n}^{p}, I_{n}^{q} \rangle\) and \(\langle J_{n}^{p}, J_{n}^{q} \rangle\) measure the similarity between video and pose pair embeddings, from two different viewpoints.

These constraints collectively enhance the geometric consistency of the learned embeddings across modalities and viewpoints.

\subsubsection{Masked Pose Modeling}
We follow the encoder-decoder design in MAE~\citep{MAE}, where the transformer encoder focuses on representation
learning while the decoder implements the reconstruction task. Our decoder takes as input the aligned pose features $\{\overline{\pi}_{1,1}^{1}, ..., \overline{\pi}_{N_{p},T}^{V}\} \in \mathbb{R}^{V \times N_{p} \times T \times d}$ output by the encoder. 
The reconstruction target corresponds to the initial coordinates of randomly sampled pose tokens that have been masked before being fed to our encoder (see Fig. \ref{fig:enter-label}). We use the $L_{Mask}$ loss as a simple mean-squared error loss between the predicted and target coordinates.

Finally, the total loss for all the pretraining objectives is defined as follows: 
\begin{equation}
L_{Align}=L_{Con}+\lambda_{1}(L_{C-Geo}+L_{I-Geo})+\lambda_{2}L_{Mask},
\end{equation}
where $\lambda_{1}$ and $\lambda_{2}$ are hyperparameters controlling the importance of the geometric consistency and masked pose prediction regularizers.

\subsection{Finetuning on Action Recognition}
Our model is trained in two stages. Following the pretraining phase described earlier, we finetune the pretrained encoders for surgical action recognition. For each modality \(M \in \{I, J\}\) and each available viewpoint \(c \in [1, \dots, C]\), we extract a global token \(CLS_{M}^{c}\). We stack these tokens and perform an average pooling operation to obtain the overall global representation. This representation is then input into a two-layer multi-layer perceptron (MLP) to generate class probabilities. This adaptable representation enables us to utilize different viewpoints in our pretraining, finetuning, and testing framework.

\section{Datasets}
\subsection{4D-OR Dataset}
\label{4D-Dataset}
The 4D-OR dataset~\citep{4DOR} encompasses $10$ simulated total knee replacement surgeries carried out in a medical simulation center under the supervision of orthopedic surgeons. The average recording duration is 11 minutes with $6,734$ images per camera. The dataset is captured from $6$ RGB-D Kinect cameras (see Table~\ref{tab1:stats_data}) strategically mounted on the OR ceiling, ensuring complete coverage of the OR. Among the $6$ available camera points of view, the sixth camera offers a different perspective. It is located on the ceiling, thus providing a quasi-bird's-eye view of the scene. The workflow in the dataset is a simulated and simplified version of an actual surgery, and the actors' roles are regularly rotated to introduce variability in the dataset. The cameras are fixed during all procedures, enabling the ablation experiments and, notably, the cross-view experiments in section~\ref{robustness_to_shift}.

\subsection{OR-AR Dataset}
\label{OR-Dataset}
The OR Activity Recognition (OR-AR) dataset~\citep{Sharghi2020} is a large-scale dataset containing close to 400 full-length videos from 85 surgical procedures (see Table \ref{tab1:stats_data}), with some lasting over 2 hours. The collection of the videos is achieved using four time-of-flight cameras positioned in critical locations on two different carts to capture the full OR comprehensively. 

\begin{figure}
    \centering
    \includegraphics[width=0.8\linewidth]{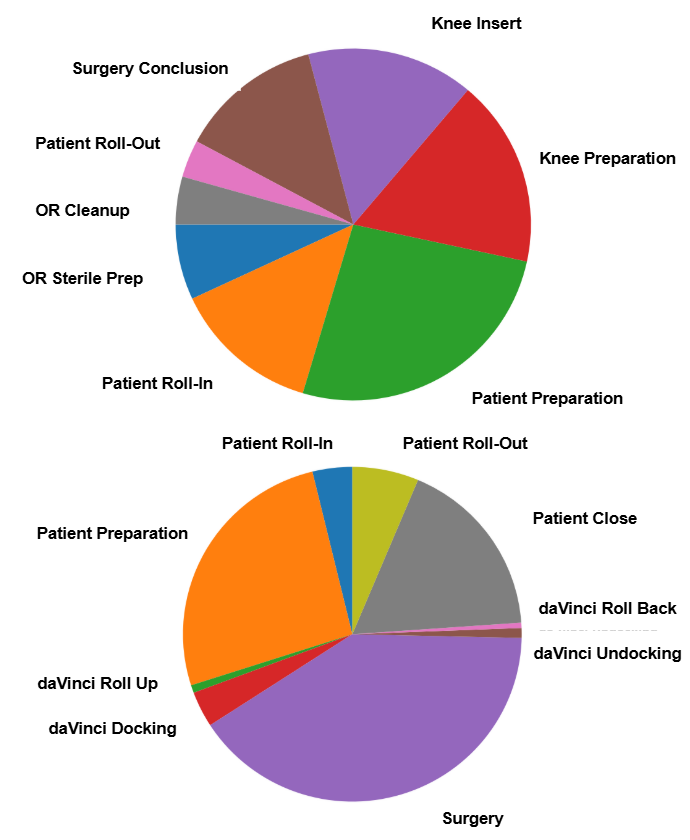}
    \caption{\textbf{Activity label distribution:} An overview of the activity durations in both 4D-OR (top) and OR-AR (bottom) datasets.}
    \label{fig:Stats}
\end{figure}

Nine activities relative to RAS are annotated on the dataset. The activity classes are highly imbalanced. The activities like \emph{Roll-back} and \emph{Roll-out} of the daVinci usually last between one and two minutes, while activities like \emph{Patient Preparation} can last more than an hour (see Fig.~\ref{fig:Stats}). The dataset includes different ORs, clinical teams, and procedures.

\begin{table}[t]
 \scriptsize

\caption{Statistics and comparison of the 4D-OR~\citep{4DOR} and OR-AR~\citep{Sharghi2020} datasets.  }\label{tab1:stats_data}
\begin{tabular}{>{\centering} m{4em} |>{\centering} m{2em} |>{\centering} m{2em} |>{\centering} m{3em} |m{5em} |>{\centering} m{4em} |m{4em}}

\rowcolor{LightCyan}
Name  & Class  & Joints 	& Modality & Frames/Case  & Cases \# & Camera \# \\
\hline
4D-OR & 8 & 17  &	 RGB  &  $\sim$700  & 10 & 6 	\\
OR-AR & 9 & 17  &	 ToF  &  $\sim$22K   & 85 & 4 	\\
\noalign{}
\end{tabular}
\end{table}

\subsection{Implementation details}
We implement our method using PyTorch~\citep{Paszke2019PyTorchAI} based on the PySlowfast library. Our baseline model is the space-time MViT-S~\citep{MViT} with MaskFeat pretraining~\citep{MaskFeat}. In all experiments, we use an input size of (8, 224, 224) and a token cube size of (2, 16, 16), which results in a total of 784 vision tokens. As mentioned in section~\ref{pose_encoder}, we set $N_{p} = 8$ as the maximum number of persons detected in a single frame for both datasets, resulting in a total of 64 pose tokens per viewpoint.

In our pretraining experiments, we exploit all available camera viewpoints in both datasets. We use a stochastic gradient descent as our optimizer, with a cosine learning rate schedule over 50 epochs, with $0.01$ as the initial learning rate and a weight decay of $1e^{-5}$. We use $\lambda_{1}=0.5$ and $\lambda_{2}=0.5$ for our pretraining experiments.

We finetune our models and baseline models using the AdamW optimizer~\citep{AdamW} with a learning rate of $1e^{-4}$ and a batch size of $16$ videos, applying cosine learning rate decay. In our experiments, we keep the weights of the first $12$ layers of the video encoder frozen.

During training, for pretraining and finetuning, images are resized to the shortest side of $256$ pixels with a random crop to $224$ pixels. For testing, images are resized to the shortest side of $224$ pixels with a center crop.

\begin{table*}[h]
\caption{Results and comparison against baselines on the 4D-OR dataset for surgical action recognition on clip classification. Accuracy (\%) is reported across three different seeds for all data fractions. }\label{tab1:ss_metrics_4D_OR}
\centering
\resizebox{0.9\textwidth}{!}{
\begin{tabular*}{1.1\textwidth}{@{\extracolsep\fill}c|c|c|cc|cccccc}
\toprule
& & &\bfseries Modalities &  & \bfseries \# Cases &     &    
  \\
\toprule
\bfseries Model  
& \bfseries PT Method &\bfseries PT Dataset &\bfseries RGB & \bfseries 2D Pose & \bfseries 1 & \bfseries 2 & \bfseries 3 & \bfseries 4 & \bfseries 5 & \bfseries 6 \\
\midrule

MViT-S & MaskFeat   & K400 & \checkmark      & \xmark          & 50.8$\pm$1.2   & 63.6$\pm$0.3  & 72.3$\pm$0.5    & 76.2$\pm$0.7   & 79.8$\pm$1.3   & 84.6$\pm$0.9       \\ 
 
ViT-B      & VMAE & K400 & \checkmark       & \xmark        & 48.8$\pm$1.2  & 62.7$\pm$0.4 & 71.5$\pm$0.7 & 74.0$\pm$1.5  & 77.3$\pm$1.0 & 81.3$\pm$0.7       \\ 
\midrule
PCT-TF &  None & N/A &  \xmark    &  \checkmark         & 38.3$\pm$2.4  & 45.2$\pm$1.1  & 51.4$\pm$0.7  & 58.0$\pm$1.2  & 54.2$\pm$1.1   & 69.5$\pm$0.8       \\ 
PCT-MViT-S  & MaskFeat     &  K400 & \checkmark    & \checkmark  &  53.2$\pm$1.5   & 64.8$\pm$0.8 & 78.5$\pm$2.4  &80.5$\pm$2.1  & 83.4$\pm$1.2   & 85.1$\pm$0.7    \\ 
\midrule
MV-CLIP  & CLIP*     & 4D-OR  & \checkmark    & \checkmark  &  54.2$\pm$1.6   & 65.9$\pm$1.5 & 79.3$\pm$1.8  & 82.8$\pm$1.4  & 84.2$\pm$1.3   & 85.5$\pm$0.7    \\ 
PreViPS & Ours    & 4D-OR & \checkmark        & \checkmark  & \bfseries 55.4$\pm$2.1  &  \bfseries 68.2$\pm$1.5 & \bfseries 80.9$\pm$2.7 & \bfseries 85.2$\pm$1.4  & \bfseries 88.7$\pm$0.8  & \bfseries 89.6$\pm$0.4        \\ 
\bottomrule
\end{tabular*}
}
\end{table*}

\begin{table*}
\caption{Results and comparison against baselines on the 
OR-AR dataset for surgical action recognition on clip classification. Accuracy (\%) is reported across three different seeds for all data fractions. }\label{tab1:ss_metrics}
\centering
\resizebox{0.9\textwidth}{!}{
\begin{tabularx}{1.1\textwidth}{@{\extracolsep\fill}c|c|c|cc|ccccc}
\toprule
& & &\bfseries Modalities &  & \bfseries Data \% & & & & \\
\toprule
\bfseries Model  
& \bfseries PT Method & \bfseries PT Dataset &\bfseries ToF & \bfseries 2D Pose & \bfseries 5\% & \bfseries 10\% & \bfseries 20\% & \bfseries 50\% & \bfseries 100\%  \\
\midrule

MViT-S & MaskFeat   & K400 & \checkmark      & \xmark          & 40.1$\pm$1.4  & 56.7$\pm$0.3  & 63.2$\pm$0.5    & 80.6$\pm$0.7   &   86.3$\pm$1.3     \\ 
ViT-B & VMAE & K400 & \checkmark       & \xmark        &   42.8$\pm$0.8 & 61.4$\pm$0.4 & 67.8$\pm$0.7 & 84.4$\pm$1.5  &  88.9$\pm$1.0     \\ 
\midrule
PCT-TF&        None & N/A & \xmark    &  \checkmark         & 38.5$\pm$1.1  & 40.7$\pm$1.1  & 51.4$\pm$0.7  & 58.0$\pm$1.2    & 64.2$\pm$1.1     \\ 

PCT-MViT-S        & MaskFeat   &  K400   & \checkmark    & \checkmark  &   45.2$\pm$0.8  & 65.6$\pm$0.8 & 70.0$\pm$2.4  &84.4$\pm$2.1   & 89.4$\pm$1.2  \\ 
\midrule
MV-CLIP        & CLIP*   &  OR-AR   & \checkmark    & \checkmark  &   54.5$\pm$1.6  & 66.4$\pm$1.3 & 71.8$\pm$2.0  & 85.2$\pm$1.2   & 90.7$\pm$1.1 \\
PreViPS & Ours    &   OR-AR   & \checkmark        & \checkmark  & \bfseries 58.4$\pm$1.3  &  \bfseries 70.1$\pm$1.5 & \bfseries 73.3$\pm$2.7 & \bfseries 87.8$\pm$1.4 &  \bfseries 92.3$\pm$0.8 \\ 
\bottomrule
\end{tabularx}
}
\end{table*}

\section{Experiments and Results}
\textbf{Baselines}: In this section, we present a series of baselines that illustrate the advantages of our pretraining strategy and our pose-centric representations.
We examine the performance of our method against five baselines: 
\begin{itemize}
 \item[$\diamond$]\textbf{MaskFeat}~\citep{MaskFeat} and \textbf{VideoMAE}~\citep{VMAE} are both state-of-the-art self-supervised approach used in general computer vision. These methods are \emph{visual-only} approaches and do not use explicit pose information. The purpose of including these methods is to demonstrate the significant improvements that can be achieved by integrating pose information into our pretraining framework. This highlights the effectiveness of both our pose architecture and our unsupervised alignment objective in utilizing multi-modal information.
\item[$\diamond$]\textbf{PCT-TF} (PCT + Transformer) method serves as a \emph{pose-only} baseline to show the performance of our pose-based approach as it has not been explored yet for the task of surgical activity recognition from the external cameras. 
\item[$\diamond$] \textbf{PCT-MViT-S} is derived from a \emph{dual-encoder architecture} without any pretraining. This architecture integrates our pose-based backbone (\emph{PCT-TF}) with the video backbone (MViT-S). The video backbone has been pretrained using the Maskfeat method on the Kinetics-400 dataset. This combination demonstrates the advantages of incorporating appearance-based features alongside our pose representations, as pose information alone may not be sufficient to identify certain surgical actions.
\item[$\diamond$]\textbf{MV-CLIP} utilizes the same architecture as PCT-MViT-S but further pretrains both encoders with a multi-view video-pose adaptation of the CLIP contrastive objective, referred to as CLIP* in section ~\ref{multi-modalContrast}.

\end{itemize}

\subsection{Data Efficient Transfer} 
\textbf{Setup}: We implement a data-efficient experimental protocol to demonstrate the advantages of our unsupervised training approach. We pretrain our \emph{MV-CLIP} and \emph{PreViPS} methods on each of the two datasets presented in sections~\ref{4D-Dataset} and ~\ref{OR-Dataset}, using all available viewpoints: $6$ for 4D-OR and $4$ for OR-AR, respectively. 
We finetune our pretrained model with progressively increasing amounts of labeled data. We use all available viewpoints for finetuning our models. We utilize the averaged global representations from every view and modality to create our clip representation.

To reduce potential bias in sampling videos from the dataset, we take each data sample three times and calculate the mean and standard deviation of the results.
For both datasets, the testing and validation datasets remain unchanged. The data-efficient performance of the models is presented in Table \ref{tab1:ss_metrics_4D_OR} and Table \ref{tab1:ss_metrics}. These tables show that our pretraining significantly improves downstream task performance. Consistently, our PreViPS model outperforms the model trained from scratch across all label percentages on both datasets. Notably, as the amount of labeled data increases, the performance gap between the best video-based baseline and our method without pretraining narrows, demonstrating the effectiveness of our pretraining strategy.



\subsection{Unimodal and Cross-View Evaluation}
Our model adapts to various input modalities and viewpoints, allowing for unimodal and cross-view setups.
The following section will explain our experimental setup for the cross-view, unimodal, and single-view experiments. We have conducted our experiments with 4D-OR because the camera setup is consistent, which helps us identify the cameras for our viewpoint ablation study.
\subsubsection{Robustness to Viewpoint Shift}
\label{robustness_to_shift}
To assess the influence of varying camera viewpoints, we conduct three experiments. In each experiment, we test on one camera viewpoint while using the other two for training. We focus on cameras 1, 4, and 6, as they provide a comprehensive overview of the scene with minimal overlap. 

As mentioned in section~\ref{4D-Dataset}, presenting the 4D-OR dataset, one view (camera viewpoint 6) gives a very different perspective of the scene and has the most minor overlap with the rest. Our results (see Table~\ref{tab:X-View}) demonstrate that alignment pretraining significantly enhances performance across all cross-view setups. This improvement is particularly pronounced when testing on the top view from camera 6, demonstrating the effectiveness of our pretraining approach.

\begin{figure*}
    \centering
    \includegraphics[width=1.0\linewidth]{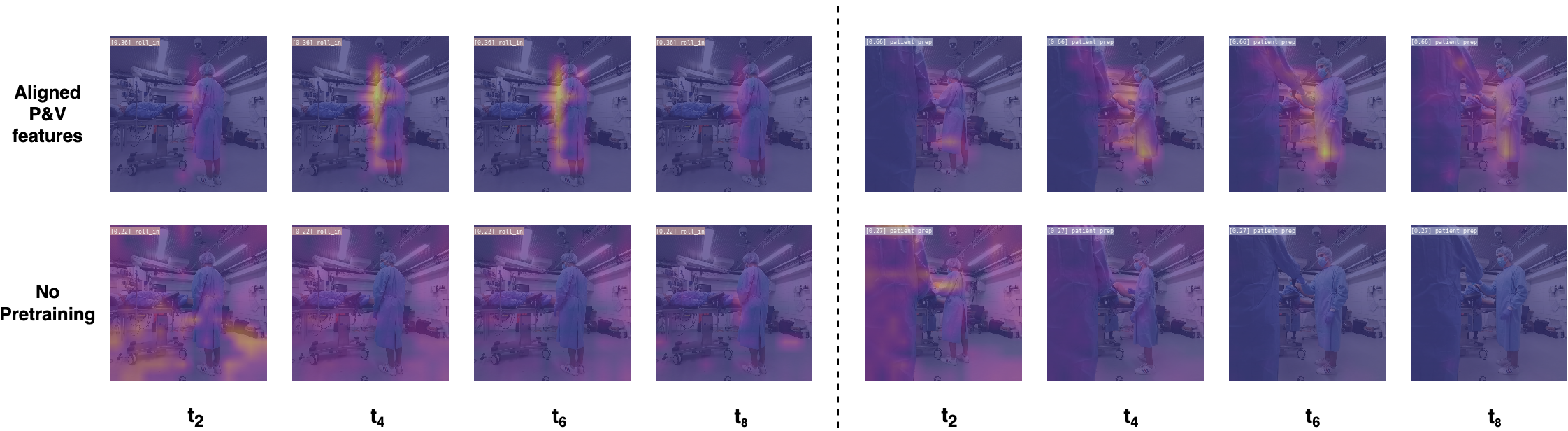}
    \caption{\textbf{GradCAM visualizations:} In the visualization of videos, brighter colors
indicate higher attention. Notably, we observe that greater attention is assigned to moving body parts. The top row shows activation maps from our pretrained model with alignment objectives, while the bottom row displays results from the model trained without video-pose alignment.}
\label{fig:visu_gradcam}
\end{figure*}

\begin{table}
    \centering
\begin{tabular}{l|c|c|c}

 \textbf{X-View Setup} & $\{1,4\};6$ & $\{1,6\};4$ & $\{4,6\};1$\\
\hline
 \emph{Accuracy Boost}  & \textbf{+6.4} & \textbf{+5.3} & \textbf{+3.5} \\

\end{tabular}
 \caption{Effectiveness of our alignment pretraining when holding out different viewpoints on 4D-OR. Performance increases are given in (\%) for different Train-Test camera setups.}
\label{tab:X-View}
\end{table}

\subsubsection{Unimodal Evaluation}
In these experiments, we investigate whether our pretraining method enhances unimodal activity recognition performance for both the \emph{pose-only} and \emph{vision-only} single modality backbones. For the pose encoder, we initialize it using weights from our multi-modal pretrained network while excluding the weights of the video encoder. Similarly, for the video backbone, we disregard the pose encoder weights.

As shown in Table~\ref{tab:UniModal}, our multi-view representation pretraining proves beneficial when finetuning on single modalities. When leveraging all available viewpoints, the pose backbone, which previously underperformed without pretraining, achieves a performance increase from $69.5\%$ to $71.9\%$. Likewise, the video backbone's performance improves modestly from $85.1\%$ to $86.8\%$. These results highlight the advantages of our pretraining approach for both modalities.

\begin{table}
    \centering
\resizebox{0.9\columnwidth}{!}{
\begin{tabular}{c|c}

 \textbf{Scratch} & \textbf{Pretrained}\\
\hline
 \textbf{(P)} \hspace{0.7cm} \textbf{(V)} & \textbf{(P)} \hspace{0.7cm} \textbf{(V)}  \\

\hline
\emph{$69.5\pm1.5$} \hspace{0.7cm} \emph{$85.1\pm0.7$} & \textbf{71.9$\pm$1.3} \hspace{0.7cm} \textbf{86.8$\pm$0.6} \\
\end{tabular}
}
 \caption{Effectiveness of our alignment pretraining when finetuning on a single modality on 4D-OR. Top-1 Accuracy is given in (\%) for both pose (P) and video (V) modalities.}
\label{tab:UniModal}
\end{table}

\subsubsection{Single-view Evaluation}
We conduct single-view experiments, where both training and testing occur from the same viewpoint. These experiments use the same viewpoints as those in the cross-view setup.  

The results in Table~\ref{tab:single_view} show a significant improvement in single-view control across different viewpoints when using our pretraining method. This highlights the benefits of multi-view representation learning, even when only a single camera is available for the downstream task.

\begin{table}
    \centering
\begin{tabular}{l|c|c|c}

 \textbf{Single View Setup} & Camera 1 & Camera 4 & Camera 6\\
\hline
 \emph{Accuracy Boost}  & \textbf{+5.6} & \textbf{+3.2} & \textbf{+5.1} \\

\end{tabular}
 \caption{Effectiveness of alignment pretraining when finetuning on a single view on 4D-OR. Performance increases are given in (\%) is given in (\%). Testing is done on the same camera viewpoint.}
\label{tab:single_view}
\end{table}

\subsection{Temporal Modeling} 
To enable in-context prediction and accurately model the sequential order of activities, we build on prior work~\citep{Sharghi2020, Jamal-ISI} and extend our finetuning approach with a recurrent neural network to capture global temporal information in the video. After extracting features from the training videos, each video is represented as \( v_{i} = \{f_{1}, ..., f_{T}\} \), where \( f_{T} \) denotes global embeddings averaged across different views and modalities. These features are then processed using a Bidirectional Gated Recurrent Unit (BiGRU)~\citep{Gru}, producing an updated feature sequence \( \overline{v_{i}} = \{\overline{f_{1}}, ..., \overline{f_{T}}\} \). The updated features are subsequently used for activity classification.  

\begin{table}
\fontsize{7pt}{7pt}\selectfont 
\centering
\caption{Results and comparison against baselines for OR surgical activity recognition on complete procedures. We provide the map, Precision, Accuracy, and F1 score.}\label{tab1:ss_metrics_temp}
\begin{tabular}{c|c|cccc}
\toprule
\bfseries Dataset
& \bfseries Model &\bfseries mAP & \bfseries Precision & \bfseries Accuracy & \bfseries F1  \\
\midrule

 & MaskFeat   &   84.2    &     82.7     & 86.0  & 80.7  \\

      & PCT-TF  &   77.1     &     75.3    & 78.3  & 75.6  \\ 

  4D-OR    & PCT-MViT-S  &   88.0     &  86.9       &  90.5 & 89.4  \\ 
      & MV-CLIP &  90.8    &  89.3    & 92.0  & 90.6 \\ 
      & PreViPS &  \bfseries 92.9    &  \bfseries 91.7    & \bfseries 94.2  &  \bfseries 93.4 \\ 
      \midrule
4D-OR & LABRADOR* & N/A & 96.0 & 97.0 & 97.0 \\
\midrule

 &   MaskFeat  & 89.5    &  87.3  &  89.8 & 88.6   \\ 
     & PCT-TF     &  75.2    &  73.6         & 77.0  & 75.6 \\ 
 OR-AR   & PCT-MViT-S      &   91.4      & 90.1  & 92.5  &  92.5 \\ 
    & MV-CLIP &  92.4    & 90.7   & 93.5  &  93.1 \\ 
    & PreViPS      & \bfseries 93.6        & \bfseries 92.0  & \bfseries 95.4  &  \bfseries 94.3 \\ 
\bottomrule
\end{tabular}
\end{table}

The results in Table~\ref{tab1:ss_metrics_temp} highlight the benefits of incorporating additional temporal modeling. The asterisk (*) in Table~\ref{tab1:ss_metrics_temp} indicates the LABRADOR baseline~\citep{Labrador}, which uses point cloud and scene graph information (using extra depth modality) and heuristic rules for activity prediction. A direct comparison is impossible since our method does not rely on semantic scene graph annotations or memory-heavy 3D point cloud data.

Instead, we present LABRADOR as an upper baseline. Our video-pose approach achieves competitive performance without requiring fine-grained scene graph supervision. 

\subsection{Ablation Study and Analysis}

We perform extensive ablation experiments to study the effect of our method's different contributions and design choices. All ablation experiments are performed on the 4D-OR dataset. 

\begin{figure}
    \centering
    \includegraphics[width=\linewidth]{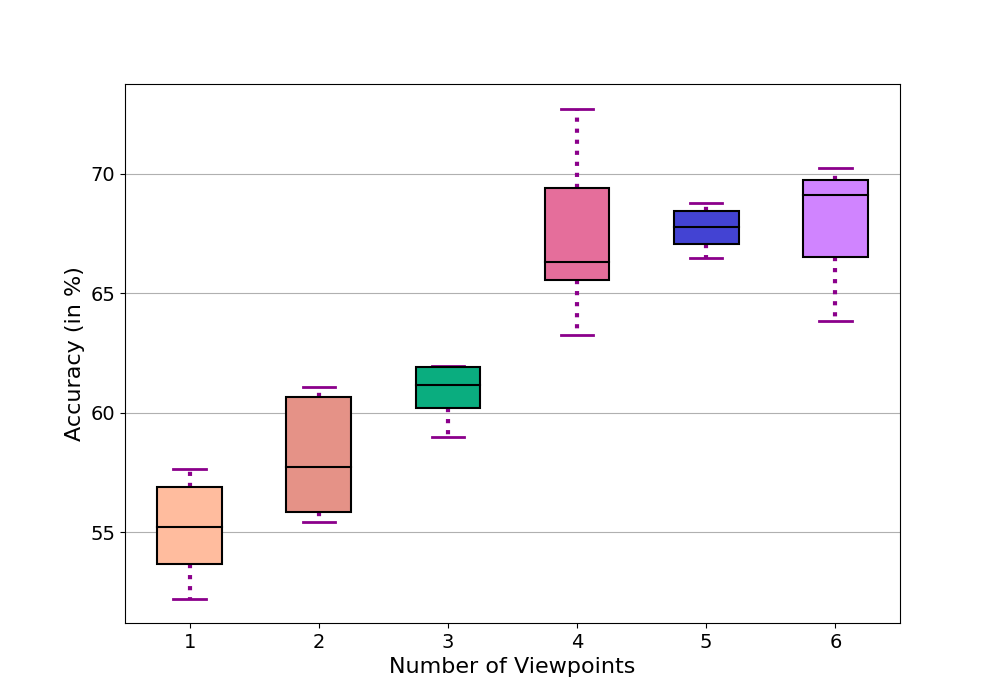}
    \caption{Box-plots showing Accuracy distributions from 4D-OR clip classification experiment for different camera viewpoints available. Ablation was run using only the pose modality as input.}
    \label{fig:abl_views}
\end{figure}

\paragraph{Effects of number of Views}
We perform an ablation study to validate
the robustness of PreViPS to varying numbers of views during inference. We
observe that the performance increases as more views are available
for representation learning. A comparison is shown in Fig. \ref{fig:abl_views} using only the pose encoder. This is intuitive as different views provide varying perspectives, which helps in recognizing actions better. 

\paragraph{Effect of pretraining objectives}
We conduct an ablation study to assess the contribution of each loss component in our pretraining objective, $L_{Align}$, as summarized in Table~\ref{tab:loss_abl}. The results show that all individual pretraining objectives are essential for achieving optimal performance.

\begin{table}
    \centering
\resizebox{0.9\columnwidth}{!}{
\begin{tabular}{c c c c}

 \textbf{MV Contrastive} & \textbf{Geometric} & \textbf{Mask Pose} & \textbf{\emph{Accuracy Drop}}\\
\hline
 \checkmark & \checkmark & \checkmark & N/A \\
 \checkmark & \xmark & \xmark & \textbf{-4.1}\\
  \checkmark & \xmark & \checkmark & \textbf{-3.3} \\
   \checkmark & \checkmark & \xmark & \textbf{-1.8} \\
\end{tabular}
}
 \caption{Effect of keeping out different unsupervised objectives on PreViPS using 4D-OR. The multi-view contrastive objective is required in our ablation study.}
\label{tab:loss_abl}
\end{table}

\paragraph{Component ablation on Pose Token Representation.} We analyze the components of our pose token representation, focusing on the choice of architecture for the pose tokenizer. Specifically, we compare a simple MLP-based tokenizer with the proposed PCT encoder. As shown in Table~\ref{tab:arch_abl}, the compositional representation of PCT leads to a significant performance improvement.  

We integrate positional embeddings with pose embeddings to accurately encode detected human poses with their timestep, track ID, and viewpoint ID. We conduct an ablation study on these positional encodings, and the results in Table~\ref{tab:arch_abl} demonstrate the performance gains achieved by incorporating them into the pose representation.  

Finally, Figure~\ref{fig:visu_gradcam} visualizes the differences in activation maps between the pretrained model and a model trained without our alignment objectives. The pretrained model shows greater attention to moving body keypoints, indicating that our multi-modal pretraining approach effectively transfers 2D pose information to the vision encoder.

\begin{table}[t!]
    \centering
\resizebox{0.7\columnwidth}{!}{
\begin{tabular}{c c}

   \textbf{Pose Token Ablation} & \textbf{\emph{Accuracy} (\%)} \\
\hline
  MLP 2D Coords & 60.3  \\
  PCT~\citep{PCT} & \textbf{69.5} \\
  \hline
   W/O Pos Embed. & 65.2  \\
    W Pos Embed. & \textbf{68.4} \\
\end{tabular}
}
 \caption{Effect of replacing the PCT~\citep{PCT} pose tokenizer with a simple MLP baseline. Benefits from adding positional embeddings in our pose token representation.}
\label{tab:arch_abl}
\end{table}


\section{Conclusion}
In this work, we present PreViPS, a novel calibration-free multi-view multi-modal pretraining framework designed for surgical activity recognition. To the best of our knowledge, PreViPS is the first framework to align 2D pose and visual embeddings across multiple camera viewpoints. This is particularly significant in the ORs where calibration is often impractical and challenging. By leveraging a tokenized discrete representation for the human pose, integrating cross- and in-modality geometric constraints, and employing masked pose modeling, our method enhances representation learning and improves surgical activity recognition performance in both multi-view and single-view settings. This work paves the way for more robust and efficient surgical activity recognition systems, capable of capturing fine-grained clinician movements and leveraging multi-view knowledge without needing calibrated camera setups or advanced point-cloud generation using RGBD cameras in complex surgical environments.


%







\section{Acknowledgements}

This work was supported by a Ph.D. fellowship from Intuitive Surgical and French state funds managed by the ANR within the National AI Chair program under Grant ANR-20-CHIA-0029-01 (Chair AI4ORSafety) and the Investments for the Future program under Grants ANR-10-IDEX-0002-02 (IdEx Unistra) and ANR-10-IAHU-02 (IHU Strasbourg). This work was granted access to the IDRIS HPC resources under the allocations 2021-AD011012367R1 and AD011011631R4 made by GENCI.
\\

\textbf{Informed consent:} Data was collected with informed consent from the human participants involved.

\bibliographystyle{model2-names.bst}
\bibliography{arxiv}

\end{document}